\pdfoutput=1

\documentclass[11pt]{article}

\usepackage[]{naacl2021}

\usepackage{times}
\usepackage{latexsym}

\usepackage{expex}

\usepackage{graphicx}
\usepackage{pgfplots}
\usepackage{tikz}
\usetikzlibrary{positioning}
\usetikzlibrary{calc}
\usetikzlibrary{shapes.multipart}

\usepackage[T1]{fontenc}

\usepackage[utf8]{inputenc}

\usepackage{microtype}

\title{Apurinã Universal Dependencies Treebank}

\author{Jack Rueter\textsuperscript{1}, Marília Fernanda Pereira de Freitas\textsuperscript{2}, Sidney da Silva Facundes\textsuperscript{2}, \\ \textbf{Mika Hämäläinen\textsuperscript{1} and Niko Partanen\textsuperscript{1}}   \\
  \textsuperscript{1}University of Helsinki \\
  \textsuperscript{2}Universidade Federal do Pará \\
  \texttt{\textsuperscript{1}firstname.lastname@helsinki.fi} \\
  \texttt{\textsuperscript{2}\{mfpf,sidi\}@ufpa.br}\\}

\begin{document}
\maketitle
\begin{abstract}
This paper presents and discusses the first Universal Dependencies treebank for the Apurinã language. The treebank contains 76 fully annotated sentences, applies 14 parts-of-speech, as well as seven augmented or new features -- some of which are unique to Apurinã. The construction of the treebank has also served as an opportunity to develop finite-state description of the language and facilitate the transfer of open-source infrastructure possibilities to an endangered language of the Amazon. 
The source materials used in the initial treebank represent fieldwork practices where not all tokens of all sentences are equally annotated. For this reason, establishing regular annotation practices for the entire Apurinã treebank is an ongoing project.  

\end{abstract}



\section{Introduction}

Apurinã (ISO code apu) is an endangered language spoken in the Amazon Basin. The language has around 2,000 native speakers and it is definitely endangered according to the UNESCO classification \cite{moseley_2010}. This paper is dedicated to describing the first ever Universal Dependencies (UD) treebank for Apurinã\footnote{https://github.com/UniversalDependencies/UD\_Apurina-UFPA}. 
We describe how the treebank was created, and what exact decisions were made in different parts of the process. 

The UD project \cite{11234/1-3424} has the goal of collecting syntactically annotated corpora containing information about lemmas, parts-of-speech, morphology and dependencies in such a fashion that the annotation conventions are shared across languages, although there may be inconsistencies between languages (see \citealt{b5a470f949d64509a1f811beadd07f6b}). 
As the number of South American languages represented in the Universal Dependencies project has grown rapidly in the last years \cite[see i.e.][]{vasquez2018toward,thomas2019universal}, the descriptions of individual treebanks are thereby also a very valuable resource that helps to maintain consistency in the treebanks of this complex linguistic regions.  

The advantage of UD treebanks is that they can be used directly in many neural NLP applications such as parsers \cite{qi2020stanza} and part-of-speech taggers \cite{kim-etal-2017-cross}. Although the endangered languages have a very different starting point in comparison with large languages \cite{mika-endangered}, there has been recent work \cite{lim2018multilingual,ens2019morphosyntactic,hamalainen2020morphological,khalid-endangered} showcasing good results on a variety of tasks even for the few endangered languages that have a UD treebank.

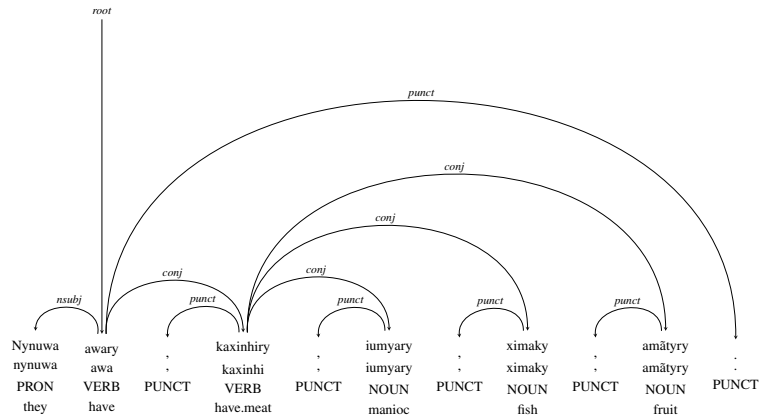
\begin{figure*}[hbtp]
\centering
\resizebox{10cm}{!}{%
\begin{tikzpicture}[
    >=stealth,
    token/.style={text height=1em, rectangle split, rectangle split parts=4},
     dep/.style={font=\small\itshape, midway, above=-.2em},
     root/.style={font=\small\itshape, above},
]
    \path[anchor=base]
        node[token] (t1) {Nynuwa\nodepart{two}nynuwa\nodepart{three}PRON\nodepart{four}they}
        node[token, right=1em of t1] (t2) {awary\nodepart{two}awa\nodepart{three}VERB \nodepart{four}have}
        node[token, right=1em of t2] (t3) {,\nodepart{two},\nodepart{three}PUNCT}
        node[token, right=1em of t3] (t4) {kaxinhiry\nodepart{two}kaxinhi\nodepart{three}VERB\nodepart{four}have.meat}
        node[token, right=1em of t4] (t5) {,\nodepart{two},\nodepart{three}PUNCT}
        node[token, right=1em of t5] (t6) {iumyary\nodepart{two}iumyary\nodepart{three}NOUN\nodepart{four}manioc}
        node[token, right=1em of t6] (t7) {,\nodepart{two},\nodepart{three}PUNCT}
        node[token, right=1em of t7] (t8) {ximaky\nodepart{two}ximaky\nodepart{three}NOUN \nodepart{four}fish}
        node[token, right=1em of t8] (t9) {,\nodepart{two},\nodepart{three}PUNCT}
        node[token, right=1em of t9] (t10) {amãtyry\nodepart{two}amãtyry\nodepart{three}NOUN\nodepart{four}fruit}
        node[token, right=1em of t10] (t11) {.\nodepart{two}.\nodepart{three}PUNCT};
    \begin{scope}[local bounding box=arcs]
        \draw[->] ($(t2.north)+(-.3em, 0)$) .. controls ($(t2.north)!0.5!-90:(t1.north)$) and ($(t1.north)!0.5!-270:(t2.north)$) .. (t1.north) node[dep] {nsubj};
        \draw[->] ($(t4.north)+(-.3em, 0)$) .. controls ($(t4.north)!0.5!-90:(t3.north)$) and ($(t3.north)!0.5!-270:(t4.north)$) .. (t3.north) node[dep] {punct};
        \draw[->] ($(t2.north)+(.3em, 0)$) .. controls ($(t2.north)!0.5!90:(t4.north)$) and ($(t4.north)!0.5!270:(t2.north)$) .. (t4.north) node[dep] {conj};
        \draw[->] ($(t6.north)+(-.3em, 0)$) .. controls ($(t6.north)!0.5!-90:(t5.north)$) and ($(t5.north)!0.5!-270:(t6.north)$) .. (t5.north) node[dep] {punct};
        \draw[->] ($(t4.north)+(.3em, 0)$) .. controls ($(t4.north)!0.5!90:(t6.north)$) and ($(t6.north)!0.5!270:(t4.north)$) .. (t6.north) node[dep] {conj};
        \draw[->] ($(t8.north)+(-.3em, 0)$) .. controls ($(t8.north)!0.5!-90:(t7.north)$) and ($(t7.north)!0.5!-270:(t8.north)$) .. (t7.north) node[dep] {punct};
        \draw[->] ($(t4.north)+(.3em, 0)$) .. controls ($(t4.north)!0.5!90:(t8.north)$) and ($(t8.north)!0.5!270:(t4.north)$) .. (t8.north) node[dep] {conj};
        \draw[->] ($(t10.north)+(-.3em, 0)$) .. controls ($(t10.north)!0.5!-90:(t9.north)$) and ($(t9.north)!0.5!-270:(t10.north)$) .. (t9.north) node[dep] {punct};
        \draw[->] ($(t4.north)+(.3em, 0)$) .. controls ($(t4.north)!0.5!90:(t10.north)$) and ($(t10.north)!0.5!270:(t4.north)$) .. (t10.north) node[dep] {conj};
        \draw[->] ($(t2.north)+(.3em, 0)$) .. controls ($(t2.north)!0.5!90:(t11.north)$) and ($(t11.north)!0.5!270:(t2.north)$) .. (t11.north) node[dep] {punct};
    \end{scope}

    \draw[->] ($(arcs.north west)!(t2.north)!(arcs.north east)$) node[root] {root} -- (t2.north);

\end{tikzpicture}
}
\caption{An example of a UD tree for an Apurinã sentence meaning \textit{`They had it, had meat, manioc, fish, fruit'}.}
\label{fig:example_tree}
\end{figure*}

The fact that UD treebanks can be used with neural models to build higher level NLP tools is one of the key motivations for us to build this resource for Apurinã. In addition to NLP research, UD treebanks have been used in many purely linguistically motivated research papers \cite{Croft2017LinguisticTM,levshina2017communicative,Levshina+2019+533+572,sinnemaki-haakana-2020-variation}. 
We believe such developments will only grow stronger, and believe that easily available treebanks in the UD project, covering continuously better the world's linguistic diversity, will continue widening their role as suitable and valuable tools for both descriptive linguistic research and computational linguistics. 
This goal will be achievable only by creating an open discussion about the conventions and choices done in different treebanks, which can be adjusted and refined at the later stage. 
This study aims to provide such description about Apurinã treebank. An example of a UD annotated sentence in Apurinã can be seen in Figure \ref{fig:example_tree}.

\section{Modelling the Apurinã Language in UD}
The Apurinã language has a rich morphology with regular correlation between numerous formatives and semantic categories. One challenge in the conversion from fieldwork/typology style annotation to that used in the UD project is to choose what features should or can be highlighted with specific transferability to other UD projects and which ones should only be represented as language specific morphology. 

The task has also been contemplated from a finite-state perspective, where regular inflection plays a decisive role in determining lemma and regular inflection strategies. Finite-state description also entails the use of the open-source GiellaLT infrastructure (Norwegian Arctic University, Tromsø) \cite{Moshagen2014}, which introduces a large number of mutual tag definitions and practices that can be applied to Apurinã with ample analogy from the morphologically challenging Uralic and other languages of the Circum-Polar region.

Solutions for dealing with the categories of case, number, person and gender are available in the GiellaLT infrastructure. 
Extensions, however, have been required for Apurinã in the categories of number, person and gender. Unlike some Indo-European and Uralic languages, the category of gender must also be applied to the subjects and objects of verbs; subject and object marking for number (see \citealt{facundes-multilingual}) and person categories could have been adapted directly from description work in the Erzya \cite{rueter2018erzud} and Moksha \cite{rueter_2018_1156112} UD treebanks.

\subsection{Case}
The Feature of \textsc{case}, for example, permeates many of the individual language projects, and some attempts are made to align case documentation with principles adapted in the Unimorph project (\citealp{kirov2018unimorph}). 
In the instance of Apurinã, parallel case categories have been adapted with names familiar to those used in work with languages of the Uralic language family. 
This was done principally because the team involved in the annotation was most familiar with this language family: at the same time the Uralic UD annotations, especially for the minority languages, are already closely adapted to the UD project at large. 
Whether such generalizations work is also one test for the cross-linguistic suitability of the current annotation model. 

The concept of case in Apurinã is most salient in oblique marking. While the subject, object and adposition complements show no special marking, there are at least six oblique marker to deal with (\citealp[385--390]{Facundes2000}). The labeling of these cases also underlines a problem not new to UD, namely, every language research tradition tends to apply its own terms for similar functions. Apurinã, as in the Uralic languages, shows evidence of case-like formatives associated not only with nominals but verbs, as well. 
In the first version of the Apurinã UD treebank, the formative case name pairs have been assigned as follows:
\textit{munhi} = Dat (dative, allative, goal), \textit{kata} = Com (comitative, associative), \textit{ã} = Loc (locative, instrumental), \textit{Ø} = Nom (nominative). Subsequent work in the dataset will introduce the additional case formative \textit{sawaky} = Temp (temporal), and show the extent of shared morphology across parts-of-speech.
\subsection{Possession}
One complexity of Apurinã morphology is encountered in the expression of possession. While the possessor of a noun may be indicated morphologically on the possessum, it is not obligatory. A preceding personal pronoun, for example, also serves as a marker of possession, to which the morphology of the possessum reacts and shows indication of being possessed. Hence, there are four basic categories that can be expressed on the possessum: person, number and gender of the possessor, on the one hand, and indication of whether the entity is a possessum or not, on the other. These categories are expressed as feature and value pairs in the UD project: 

\begin{itemize}
\setlength\itemsep{0em}
  \item Gender[psor]=Masc|Fem
  \item Number[psor]=Plur|Sing
  \item Person[psor]=1|2|3
  \item Possessed=Yes|No
\end{itemize}

While matters of gender, number and person are directly attested in the morphology of the possessum, the feature \textsc{Possessed} identifies the individual noun as to whether there is or is not marking indicating that it is possessed. This particular issue of research is dealt with extensively in  \citealp{Freitas2017}.

Apurinã nouns can be split into four groups on the basis of how their morphology is affected by possession. 
There are nouns that never take possession or possessive affixes. Such nouns include proper names (\citealp[179--180]{Freitas2017}). The remaining nouns, however, take possessive affixes, on the one hand, and additional marking to indicate whether the word is possessed or not. First, 
there are nouns, such as kinship terms, that virtually always appear with possessive affixes and no morphology to indicate that they are possessed. These nouns may only be construed as not possessed in some verbal incorporations where the noun is non-specific by nature. A formative \textit{-txi} is present to indicate the noun is not possessed. Other words in this group, including terms for body parts and individual belongings, for example, take the \textit{-txi} formative to indicate the item is not possessed more freely, e.g. \textit{kywy} 'head (possessed)' vs \textit{kywĩtxi} 'head (possessed)' (\citealp[163-171]{Freitas2017}; \citealp[199-204,228-236]{Facundes2000}). Second, there are noun categories that take the formatives \textit{-ne}, \textit{-te} and \textit{-re1} to indicate the item is possessed, but they, in contrast, have no morphology to indicate that the item is not possessed. Third, there is group of nouns which actually mark both the possessed with the formative \textit{-re2} and the non-possessed with the formative \textit{-ry2}. This alternation is described in \citealp{Facundes2000}, and explicitly \citealp[(112-123)]{Freitas2017} (see Table~\ref{tab:possessed-marking})

\begin{table}[]
\centering
\small
\begin{tabular}{|l|l|l|l|}
\hline
    & Possessed                                    & Not Possessed &   translation                                         \\ \hline
body part & kywy  & kywĩ-txi  &  'head'     \\ \hline
person & sytu-re  & sytu &  'woman'     \\ \hline
other & kuta-re2  & kuta-ry2 & 'basket'     \\ \hline
\end{tabular}
\caption{Marking of possessed feature}
\label{tab:possessed-marking}
\end{table}


The Apurinã treebank solution has been to introduce the \textbf{possessed} feature with \textbf{Yes} and \textbf{No} values. Nouns that cannot be possessed are simply left without the feature Possessed. 

\subsection{Intransitive descriptive verbs}
Apurinã verbs can bear morphology indicating subject and object, be that simultaneously or separately. What is interesting, however, is that a specific subclass of intransitive descriptive verbs attest to the use of object marking to indicate congruence with the subject 
(\citealp[278–283]{Facundes2000}). There are, in fact, certain verbs that distinguish object and subject marking strategies for the same intransitive verbs, such that subject marking indicates a short temporal frame, and object marking indicates permanency (cf. \citealp{Chagas2007}; \citealp[70--71]{Freitas2017}).

The solution here has been to refer to object-looking morphology with subject congruence as subject marking:

\begin{itemize}
\setlength\itemsep{0em}
  \item Gender[subj]=Fem|Masc
  \item Number[subj]=Plur|Sing
  \item Person[subj]=1|2|3
\end{itemize}

To cope, an additional feature value set has been introduced to distinguish verbs of the intransitive descriptive (Vid) nature, and this subset is subsequently split on the on basis of whether the formative entails object-identical \textit{Vido} or subject-identical marking \textit{Vids}.


\subsection{Derivations}
Fieldwork annotations of certain derivational morphology are minimalistic, and their conversion in the UD treebank calls for more specific representation. Whereas some formatives have been referred to using the same terms, e.g. nominalizer, gerund, we have been obliged to elaborate. 
Only one feature has been provided for Derivation, Proprietive (\textit{ka-}). The proprietive construction is one of many annotated as \textbf{atrib} in the fieldwork materials. 

\subsection{Lemmatization}

    The Apurinã language is spoken in 18 indigenous communities of the Purus basin (\citealp{limaEtAl2019}). Grammar descriptions from \citealp{Facundes2000} to \citealp{Freitas2017} demonstrate a change in orthographic development, on the one hand, and actual variation in forms of the same words in relation to geographic location, on the other. Materials in the treebank alone show some vacillation with regard to stem-initial \textit{h} and word-internal \textit{e} vs \textit{i}. Since the orthographic standard is still in a developmental state, lemma forms have been chosen on a basis of whether they occur in the manuscript dictionary (\citealp{Lima-Padovani2016}) or not, and a preference for longer word forms, i.e., \textit{h}-initial stems are forwarded, since it easier to drop a letter in the description than to automatically insert one. Thus the form \textit{hãty} 'one' is given as a lemma instead of its variant \textit{ãty} (as given in the dictionary), and \textit{herãkatxi} (given as a variant) is forwarded as a lemma over both \textit{erãkatxi} and \textit{er\~{e}katxi} (given in the examples of the alphabet), \textit{ar\~{e}katxi}.
    The high vowel \textit{i} is preferred over the middle \textit{e} such that \textit{tiwitxi} 'thing' is given as a lemma for the forms \textit{teetxi} and \textit{tiitxi}. Fortunately, work with Apurinã variation is continuing (\citealp{limaEtAl2019}), and an updated version of the Apurinã-Portuguese dictionary is forthcoming.

\section{Treebanks in figures}
There were 76 valid and dependency-annotated sentences in the first release. Broken into figures, these sentences contain 574 tokens and a 454 word count, which can be further broken down into features, parts-of-speech and dependency relations.

The most salient features are \textit{Case} (101), \textit{Gender} (96),
\textit{Number} (73), but the newly introduced \textit{Gender[obj]} (47) is also well attested. The \textit{Case} feature owes its prominence to the presence of all nouns not marked for oblique cases, i.e. \textit{Nom}; this leaves a total of 25 obliques (see Table~\ref{tab:ud-features}).

\begin{table}[h]
\centering
\tiny
\begin{tabular}{|l|l|l|l|}
\hline
Feature    & №    &  Feature    & №        \\ \hline
AdvType=Tim & 1 & Number[obj]=Plur,Sing & 1 \\ \hline
Aspect=Prog & 1 & Number[obj]=Sing & 51 \\ \hline
Case=Com & 4 & Number[psor]=Sing & 10 \\ \hline
Case=Dat & 7 & Number[subj]=Plur & 1 \\ \hline
Case=Loc & 11 & Number[subj]=Sing & 7 \\ \hline
Case=Nom & 76 & Person=3 & 53 \\ \hline
Case=Temp & 3 & Person[obj]=3 & 52 \\ \hline
Derivation=Proprietive & 2 & Person[psor]=3 & 8 \\ \hline
Gender=Fem & 14 & Person[subj]=3 & 8 \\ \hline
Gender=Masc & 82 & Possessed=No & 27 \\ \hline
Gender[obj]=Masc & 47 & Possessed=Yes & 8 \\ \hline
Gender[psor]=Fem & 3 & PronType=Prs & 53 \\ \hline
Gender[psor]=Masc & 11 & VerbForm=Conv & 2 \\ \hline
Gender[subj]=Masc & 8 & VerbForm=Vnoun & 9 \\ \hline
Number=Plur & 16 & VerbType=Vido & 2 \\ \hline
Number=Sing & 57 & & \\ \hline

\end{tabular}
\caption{\label{tab:ud-features}Features}
\end{table}

The most prominent parts-of-speech the \textsc{noun} (170) and \textsc{verb} (137) classes, followed by \textsc{pron} (59) and \textsc{adv} (39), whereas two instances of the same unknown word \textit{pekana} outnumber the \textsc{adj}, \textsc{cconj} and \textsc{propn}, each at one (see Table~\ref{tab:ud-posfigures}). 
\begin{table}[h]
\centering
\small
\begin{tabular}{|l|l|l|l|l|l|}
\hline
PoS  & №   & PoS  & № & PoS  & № \\ \hline
ADJ   & 1   &DET   & 11  & PROPN & 1  \\ \hline
ADP   & 3   &NOUN  & 170 & SCONJ & 3  \\ \hline
ADV   & 39  & NUM   & 9   & VERB  & 137  \\ \hline
AUX   & 6    & PART  & 13    & X     & 2 \\ \hline
CCONJ & 1      & PRON  & 59  & & \\ \hline

\end{tabular}
\caption{Part-of-speech Figures}
\label{tab:ud-posfigures}
\end{table}

An important dependency relation (\textit{deprel}) is \textit{nsubj} (83), which is made possible through the extensive use of the \textit{conj} relation. 
Language-specific \textit{deprels} have extensions such as:  
\textit{lmod} = locative modifier,
\textit{neg} = negation,
\textit{poss} = possession,
\textit{relcl} = relative clause
\textit{tcl} = temporal clause and
\textit{tmod} = temporal modifier (see Table~\ref{tab:ud-deprel}).

\begin{table}[h]
\centering
\small
\begin{tabular}{|l|l|l|l|l|l|}
\hline
deprel & № &  deprel  & № &  deprel    & № \\ \hline
acl & 10 & mark & 3 & advmod:lmod & 1  \\ \hline
advcl & 5 & nmod & 18 & advmod:neg & 13 \\ \hline
advmod & 22 & nsubj & 83 & advmod:tmod & 13 \\ \hline
aux & 5 & nummod & 9 & nmod:poss & 2 \\ \hline
case & 3 & obj & 63 & nsubj:cop & 2 \\ \hline
cc & 3 & obl & 15 & obj:agent & 1 \\ \hline
conj & 48 & root & 76 & obl:lmod & 19 \\ \hline
dep & 2 & xcomp & 1 & obl:tmod & 4 \\ \hline
det & 24 & acl:relcl & 5 & & \\ \hline
csubj & 2 & advcl:tcl & 2 & & \\ \hline

\end{tabular}
\caption{Dependency relations}
\label{tab:ud-deprel}
\end{table}

\section{Future work}

Due to the size and orientation of the dataset some features of the Apurinã language have been neglected. 
It will also be a challenge to apply recent studies in 
noun incorporation annotation for UD in \citealp{tyers2020dependency} to what \citealp[]{Facundes2015} describe for Apurinã noun and classifier incorporation.







Another obvious goal for further work is to make Apurinã treebank so large that it can be split into train, test and dev portions. 
The goal to expand the treebank is connected to the availability of resources. 
Currently the sentences used in the treebank come mainly from the grammatical descriptions. 
As a language documentation corpus exists\footnote{\url{https://elar.soas.ac.uk/Collection/MPI1029704}}, an important consideration is whether the treebank sentences could be more closely connected to audio and video recordings as well, and, of course, the main corpora in Belém, as multimodal resources are valuable in language documentation. 


\bibliography{anthology,custom}
\bibliographystyle{acl_natbib}

\end{document}